%% file: main.tex
\pdfoutput=1
\documentclass[11pt]{article}

\usepackage[preprint]{acl}
\usepackage{times}
\usepackage{latexsym}
\usepackage[T1]{fontenc}
\usepackage[utf8]{inputenc}
\usepackage{microtype}
\usepackage{inconsolata}

\usepackage{amsmath}
\usepackage{booktabs}
\usepackage{graphicx}
\usepackage{float}
\usepackage{multirow}
\usepackage{xcolor}

\usepackage{bbding}

\definecolor{DropRed}{RGB}{185, 42, 42}
\definecolor{NearTeal}{RGB}{0, 128, 128}
\definecolor{BiasOrange}{RGB}{204, 102, 0}
\newcommand{\sysname}{\textsc{ReKey}}
\newcommand{\drop}[1]{\textcolor{DropRed}{$\downarrow$\,#1}}
\newcommand{\near}[1]{\textcolor{NearTeal}{#1}}

\definecolor{CotBlue}{RGB}{70, 100, 170}

\newcommand{\modelname}[2]{%
  \raisebox{-0.15em}{\includegraphics[height=1.2em]{#1}}%
  \hspace{0.45em}#2%
}

\usepackage{todonotes,xspace}

\title{\sysname{}: Metadata-Grounded Visual-Key Regeneration for Contamination-Resilient VQA Evaluation}

\author{
 \textbf{Tengjie Lin\textsuperscript{1,*}},
 \textbf{Yutao Sun\textsuperscript{1,*}},
 \textbf{Jingwei Ni\textsuperscript{2,*}},
 \textbf{Shuhan Ge\textsuperscript{3}},\\
 \textbf{Hao-Xuan Ma\textsuperscript{4}},
 \textbf{Yanting Miao\textsuperscript{5}},
 \textbf{Wangyue Lu\textsuperscript{1}},
 \textbf{Mingshuai Chen\textsuperscript{1}},\\
 \textbf{Tiancheng Zhao\textsuperscript{6}},
 \textbf{Jianwei Yin\textsuperscript{1}}
\\
\\
 \textsuperscript{1}Zhejiang University,
 \textsuperscript{2}ETH Z\"urich,
 \textsuperscript{3}Nanjing University of Science and Technology,
\\
 \textsuperscript{4}Nanjing University,
 \textsuperscript{5}University of Waterloo,
 \textsuperscript{6}Binjiang Institute of Zhejiang University
\\
 \small{
   \textsuperscript{*}Equal contribution. \textbf{Correspondence:} \href{mailto:m.chen@zju.edu.cn}{m.chen@zju.edu.cn}, \href{mailto:tianchez@zju-bj.com}{tianchez@zju-bj.com}, \href{mailto:zjuyjw@zju.edu.cn}{zjuyjw@zju.edu.cn}
 }
}

\begin{document}
\maketitle

\input{sections/abstract}
\input{sections/introduction}
\input{sections/related_work}
\input{sections/method}
\input{sections/experiments}
\input{sections/conclusion}
\input{sections/limitations}

\bibliography{references}

\appendix
\input{appendices/appendix}

\end{document}

%% file: sections/abstract.tex
\begin{abstract}
Static visual question answering (VQA) benchmarks age quickly: Once the items leak into training corpora, scores can reflect memorization rather than genuine visual ability, thus obscuring real progress. Rebuilding high-quality benchmarks such as V*Bench~\cite{wu2024vstar} requires substantial human annotation, yet each static release can quickly become another leaked artifact. We propose \sysname{}, a live benchmark protocol that randomly regenerates the answer-bearing local detail, or visual key, in real images at evaluation time. Using human-validated edit slots, \sysname{} samples fresh instances with new answers, construction-grounded labels, and controlled visual-search difficulty. On V*Bench, the \sysname{} regenerated benchmark reveals a sharp score jump across eight frontier vision-language models (VLMs): The original items score 9.5--18.8 percentage points higher than the regenerated variants. By making the visual key renewable, \sysname{} keeps evaluation fresh as
models and training data evolve.\footnote{Code: \url{https://github.com/Xxxxxsun/rekey-vstar}; Benchmark: \url{https://huggingface.co/datasets/Xfgll/rekey-vstar-benchmark}}
\end{abstract}

%% file: sections/introduction.tex
\section{Introduction}
\label{sec:introduction}

Static visual question answering (VQA) benchmarks become less reliable once their underlying items are exposed through training corpora or repeated development cycles. In such cases, benchmark scores may conflate visual reasoning with memorization, making it hard to determine whether a model is genuinely grounding its answer in the image or merely recalling previously encountered associations. This issue is particularly pronounced when the correct label depends on a subtle visual detail that is difficult to locate.
%

\begin{figure}[t]
  \centering
  \includegraphics[width=\columnwidth]{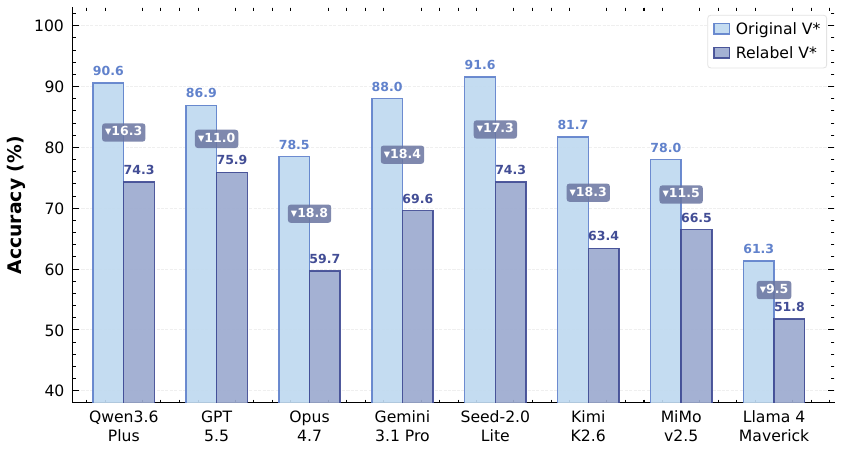}
  \caption{Across eight evaluated models, accuracy decreases by 9.5--18.8 percentage points when the same images are paired with fresh human-written questions targeting different visual keys, indicating that the original V*Bench scores are partially inflated by memorized answer-bearing details (see details in Appendix~\ref{app:contamination}).}
  \label{fig:contamination}
\end{figure}

V*Bench \cite{wu2024vstar} makes this failure mode visible.
Each of its 191 questions hinges on identifying a \emph{visual key}: a small, peripheral detail in the image that determines the correct answer.
When the same images are paired with freshly constructed questions targeting different visual keys, the performance of all evaluated models drops by 9.5--18.8 percentage points; see Figure~\ref{fig:contamination}.
The magnitude of this decline suggests that the original scores capture not only visual search capability, but also gains from memorized answer-bearing details.


Existing dynamic evaluation methods address aspects of this problem, but none satisfies the three key requirements simultaneously: (i) preserving the original
visual-search difficulty, (ii) renewing leaked answer, and (iii) maintaining label reliability without a model in the verification loop.
Approaches that source new content, whether from recent publications
\cite{shabtay2025livexiv,white2025livebench} or from
text-to-image synthesis \cite{bao2024autobenchv}, can produce fresh answers
but sacrifice control over whether each item matches the original
visual difficulty.
Methods that retain the benchmark structure, such as Vision-Language
Bootstrapping \cite{yang2025vlb}, better maintain difficulty but often preserve the
original answer -- precisely the information that contaminated models may have memorized.
Parametric regeneration further demonstrates that answers can be derived from the
construction process \cite{zou2025dynamath}, but existing examples are
largely confined to mathematical or diagrammatic settings.

For natural-image VQA, jointly addressing these gaps induces three
constraints:
(1)~\emph{Preserve difficulty}: The regenerated key should remain a small, plausible target within the original cluttered scene, rather than becoming a salient insertion or a synthetic substitute for the image.
(2)~\emph{Renew the answer}: Each benchmark instance should compose its answer from newly sampled visual evidence, rather than reusing leaked object-attribute associations or question-answer pairs.
(3)~\emph{Keep labels reliable}: The correct answer should follow from the construction process itself, since a vision-language model (VLM) tasked with inspecting small edited regions can miss relevant details or inherit the same visual biases the evaluation seeks to measure.

We address these constraints with \sysname{}, short for Metadata-Grounded Visual-Key Regeneration.
\sysname{} treats the visual key -- rather than the entire VQA item -- as the unit of renewal.
We instantiate the framework on V*Bench, which exemplifies the
small-target visual-search setting where contamination poses a particularly severe
threat to evaluation validity.
Human annotators mark edit slots in the V*Bench images once. At evaluation time, a rule-based pipeline samples fresh visual content, applies a localized edit, and derives the ground-truth label from the same sampling record that produced the edit.
This design allows a single set of annotations to support multiple fresh benchmark instances, each paired with a different correct answer grounded in the construction process.

Our main contributions are as follows.
\begin{itemize}
    \item \textbf{Visual-key regeneration.}
    We introduce visual-key regeneration as a contamination-resilient
    evaluation principle for VQA: It renews the specific detail that
    determines the answer while preserving the visual-search difficulty, turning each source image into multiple fresh instances with distinct correct answers.

    \item \textbf{Metadata-grounded answer construction.}
    We propose a construction procedure that jointly generates the edited image
    and its ground-truth label, ensuring that answers are fully determined by the construction metadata rather than model-based judgment.

    \item \textbf{Comprehensive V*Bench evaluation.}
    By pairing the same images with fresh human-written questions, we show that original V*Bench scores across eight VLMs are inflated by 9.5--18.8 percentage points. \sysname{} closes this gap while
    preserving the original difficulty and consistent scores. We further demonstrate the importance of human curation over VLM self-annotation in maintaining label reliability.
\end{itemize}

Overall, \sysname{} shows that contamination resistance and evaluation fidelity are not inherently at odds: By renewing only the visual key that determines the answer, a fixed set of source images can indefinitely support fresh, comparable evaluation.

%% file: sections/related_work.tex
\section{Related Work}
\label{sec:related}

\paragraph{Multimodal evaluation benchmarks.}
Multimodal evaluation has moved from broad capability suites to benchmarks
that stress fine-grained evidence use. V*Bench \cite{wu2024vstar} frames visual search as a core VLM capability,
with each question requiring the model to locate a visual key in a cluttered
image.
Recent benchmarks extend this pressure to high-resolution and dense scenes
\cite{zhang2025mmerealworld,zhang2025hrscene,gavrikov2025visualoverload}.
These benchmarks motivate our setting, but they remain static once their
fixed image--question--answer items are exposed.

\paragraph{Data contamination and dynamic evaluation.}
Data contamination has become a central concern for multimodal evaluation:
recent studies show leakage through both image and text channels
\cite{DBLP:conf/emnlp/SongLWCSW25}, and show that models can exploit non-visual shortcuts
when benchmark items are exposed \cite{DBLP:journals/corr/abs-2511-04655}. A recent survey
therefore treats dynamic evaluation as a central response to contamination
\cite{chen2025recentadvanceslargelangauge}. One line uses temporal freshness: LiveBench
\cite{white2025livebench} and LiveXiv \cite{shabtay2025livexiv}
collect new items from recent sources. Another line
uses regeneration or transformation: DynaMath \cite{zou2025dynamath}
instantiates mathematical VQA problems from programs, while Vision-Language
Bootstrapping \cite{yang2025vlb} rebuilds image-question pairs through visual
and language-side transformations. Together, these methods show three useful
directions: sourcing unseen content, varying existing items, and constructing
answers from generation records. Scarce-visual-key evaluation requires these
directions to meet at a more specific point: the benchmark must keep the
original visual-search problem while replacing the small piece of evidence
that determines the answer.

\paragraph{Automatic benchmark construction and verification.}
Structured VQA provides a long-standing basis for reliable benchmark
construction. CLEVR \cite{johnson2017clevr} and GQA \cite{hudson2019gqa}
derive answers from structured scene representations, making labels explicit
rather than post-hoc. Recent work further automates benchmark construction
with models: AutoBench-V \cite{bao2024autobenchv} uses text-to-image
generation and VLM orchestration to create on-demand VQA evaluations, while
KBE-DME \cite{zhang2025kbedme} evolves static VQA samples by re-selecting
visual information or adding external knowledge. Such pipelines scale
benchmark renewal, but scarce-visual-key evaluation adds a specific
constraint: the construction process must not move the evidence toward
salient or easy regions, and the label should not depend on a model judge that
may inherit evaluation-time biases. Recent VLM-as-a-judge work reports
modality neglect, instability, and systematic preferences under controlled
perturbations \cite{lee2026mmjudgebias}. \sysname{} therefore uses human
annotation to identify where a difficult visual key can exist, then uses
rule-based sampling and metadata-grounded answer construction to instantiate
fresh evidence.

%% file: sections/method.tex
\section{Method}
\label{sec:method}

\begin{figure*}[t]
  \centering
  \includegraphics[page=1,width=\textwidth]{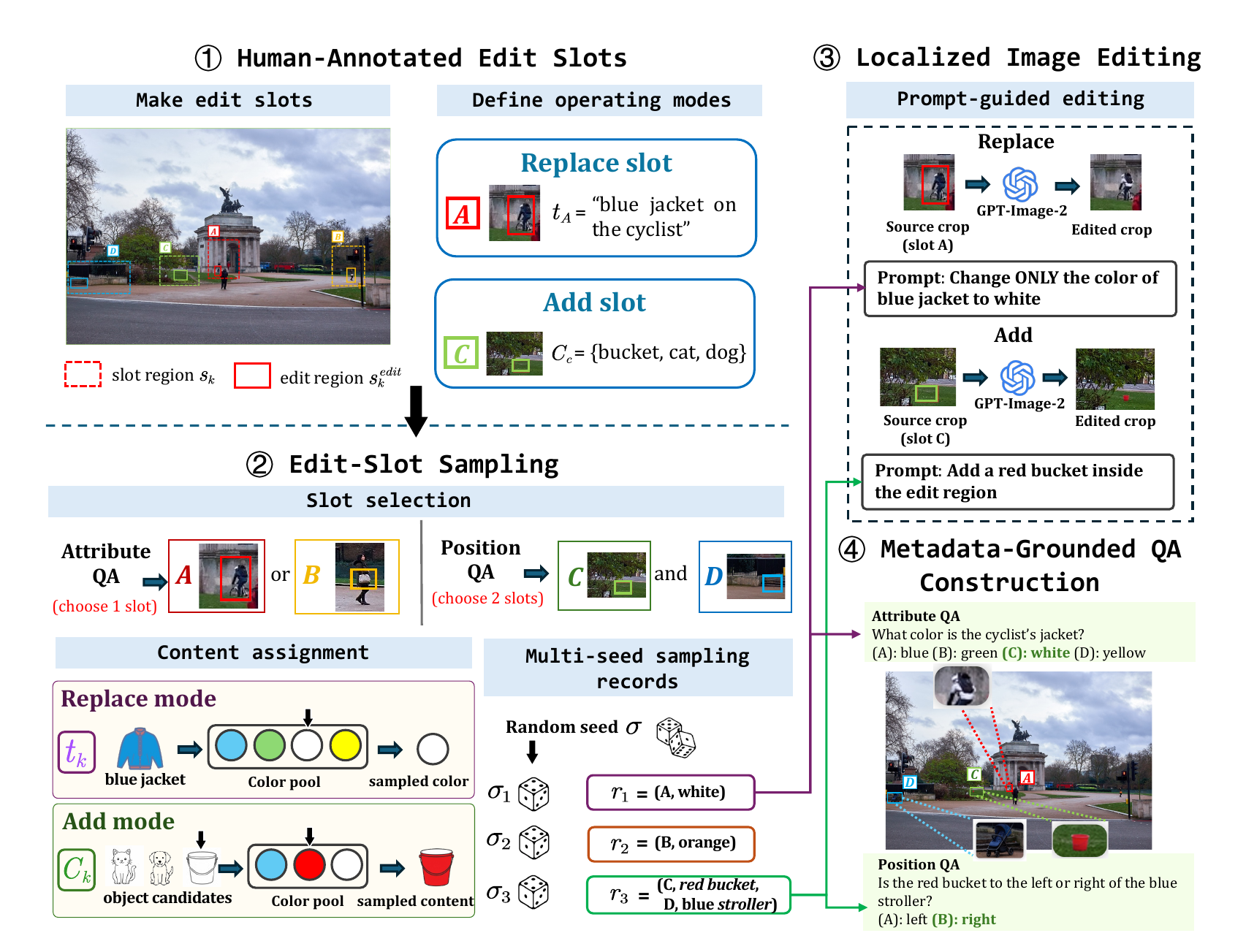}
  \caption{Overview of \sysname{} on one V*Bench source image with four annotated edit slots: replace slots A (blue jacket) and B (pink handbag), add slots C (lawn) and D (beside stairs). Three seeds produce three records: $r_1{=}(\text{A},\, \textit{white})$, $r_2{=}(\text{B},\, \textit{orange})$, $r_3{=}(\text{C},\, \textit{red bucket},\; \text{D},\, \textit{blue stroller})$. Each $r_i$ determines both the localized edit to $I$ and the fresh $(Q', A')$ pair.}
  \label{fig:overview}
\end{figure*}

Refreshing a VQA item means changing the detail that determines the answer
without changing how hard it is to find.
Given a source image $I$, question $Q$, and answer $A$, \sysname{} produces a
fresh item $(I', Q', A')$ with a different visual key but comparable visual-search
difficulty,
by separating a one-time annotation step from a fully automated
evaluation-time pipeline.
Human annotators examine each V*Bench source image and mark edit slots
$\{s_1, \ldots, s_K\}$, compact regions where a visual key can be renewed
while remaining natural and non-obvious (\S\ref{sec:annotation}).
At evaluation time, a \emph{sampling record} $r$ selects a slot and assigns
it fresh visual content (\S\ref{sec:sampling}); a localized edit produces
$I'$ from $I$ (\S\ref{sec:generation}), and a rule-based constructor derives
$Q'$ and $A'$ directly from $r$ (\S\ref{sec:qa-construction}).
Because the edit and the label are both determined by $r$, the ground truth
is fixed before any model sees the result.

\subsection{Human-Annotated Edit Slots}
\label{sec:annotation}

Each edit slot $s_k$ is an annotated region of the source image that
specifies where a visual key may be renewed without changing the
visual-search difficulty of the original item.
Within $s_k$, a smaller edit region $s_k^\text{edit}$ marks the pixels that
may change; the surrounding area in $s_k$ provides the image editor with
scene context.
Each slot operates in one of two modes, differing in how the object
candidate is determined.
A \emph{replace} slot targets a single existing object identified by a
description $t_k$.
An \emph{add} slot provides multiple object candidates in a per-slot set
$\mathcal{C}_k$.
The sampler later selects content and assigns visual attributes for each
slot (\S\ref{sec:sampling}).

Selecting where to place a slot requires human judgment: the location must
be plausible in the scene, unambiguous to reference, and difficult relative
to surrounding clutter.
Model-selected regions can drift toward central or salient objects, as
observed in VLB's GPT-4V-guided region selection \cite{yang2025vlb}; such
regions would make the refreshed item easier than the original small-target
task.
Human annotators instead mark compact locations where a new or modified key
can remain natural and non-obvious.

In \emph{replace} mode, the visual key already exists in the source image;
its visual attribute will be modified while the object identity, location,
and surrounding clutter remain the same, preserving the search difficulty by
construction.
$t_k$ is needed because $s_k$ may crop only a small or ambiguous region: it
tells the editor which object to modify, and provides a stable referent for
$Q'$ afterward.
In Figure~\ref{fig:overview},
$t_\text{A} = \text{``blue jacket on the cyclist''}$ identifies the jacket
as the target within $s_\text{A}$.

In \emph{add} mode, no existing object serves as the key; instead, a new
object will be inserted at an empty region.
The annotator curates $\mathcal{C}_k$ with objects that are plausible in
the local scene, absent from the source image, and small enough to preserve
the search challenge.
In Figure~\ref{fig:overview}, $\mathcal{C}_\text{C}$ lists objects such as
a bucket, a cat, or a dog for the lawn patch, while
$\mathcal{C}_\text{D}$ lists a stroller, a skateboard, or a bottle for the
space beside the stairs.

Together, the annotated slots encode where and how a visual key can be
renewed. This annotation is a one-time effort: once the slots are marked,
the pipeline can generate fresh items indefinitely while each inherits the
difficulty and naturalness constraints that the annotator established.

\subsection{Edit-Slot Sampling}
\label{sec:sampling}

The annotated slots define what can change; sampling decides what actually
changes in each benchmark instance.
At evaluation time, a random seed $\sigma$ determines a sampling record $r$
through two steps: slot selection and content assignment.

\sysname{} supports two question types, and the question type determines how
many slots are selected.
An \emph{attribute} question ($Q_\text{attr}$) asks about a single object's
property, so it selects one slot, yielding $r = (s, c)$ where $s$ is the
slot and $c$ the sampled content.
A \emph{position} question ($Q_\text{pos}$) asks about the spatial relation
between two objects, so it selects a pair, yielding
$r = (s_j, c_j,\; s_k, c_k)$; the sampler ensures the two visual keys are
distinct.

For each selected slot, content is assigned according to its mode: a replace
slot keeps its existing object and draws a new color from the shared pool;
an add slot first draws an object from $\mathcal{C}_k$, then draws a color
from the same pool.
In Figure~\ref{fig:overview}, three seeds produce three different records:
$r_1 = (\text{A},\, \textit{white})$ recolors the blue jacket
($t_\text{A}$) to white;
$r_2 = (\text{B},\, \textit{orange})$ recolors the pink handbag
($t_\text{B}$) to orange;
$r_3 = (\text{C},\, \textit{red bucket},\; \text{D},\,
\textit{blue stroller})$ selects add slots C and D with their respective
objects and colors.

Slot selection and content assignment together determine the pool of fresh
items derivable from one source image.
Let $|\mathcal{A}|$ denote the color pool size and $n_k$ the content choices
per slot ($|\mathcal{A}|$ for replace,
$|\mathcal{C}_k| \cdot |\mathcal{A}|$ for add).
The number of distinct attribute items is $\sum_k n_k$ and the number of
distinct position items is $\sum_{j<k} n_j \cdot n_k$, so the pool grows
combinatorially (practical constraints such as object uniqueness and
spatial separation reduce the pool slightly;
Appendix~\ref{app:construction}). Each seed thus produces a fresh
$(I', Q', A')$ with a
different correct answer, all from the same one-time annotations.

\subsection{Localized Image Editing}
\label{sec:generation}

The sampling record specifies what should change; the image editor realizes
it as a localized edit in the source image.
We use GPT-Image-2 \cite{openai2026gptimage} as the editing backbone.
The pipeline processes each selected slot through three stages:
\emph{cropping}, \emph{prompt-guided editing}, and \emph{compositing}.

\emph{Cropping.}
For each selected slot $s$, the pipeline crops $s$ from $I$ and resizes it
to a standard canvas.
The full slot region provides scene context, but the editor is instructed to
modify only the edit sub-region $s^\text{edit}$.

\emph{Prompt-guided editing.}
A rule-based prompt generator constructs an instruction from $r = (s, c)$
and $t_s$ by filling a template:

\vspace{2pt}
\noindent
\resizebox{\columnwidth}{!}{%
\small
\begin{tabular}{@{}lp{0.76\columnwidth}@{}}
\toprule
\textbf{Mode} & \textbf{Core instruction template} \\
\midrule
Replace & ``Change ONLY the color of \{$t_s$\} to \{$c$\}'' \\
Add & ``Add a \{$c$\} in the marked area'' \\
\bottomrule
\end{tabular}%
}
\vspace{2pt}

\noindent $t_s$ and $c$ fully determine the instruction; no prompt is
manually written.
In Figure~\ref{fig:overview}, $r_1$ fills the replace template as
``Change ONLY the color of blue jacket to white'';
$r_3$ fills the add template for slot C as ``Add a red bucket
in the marked area.''

\emph{Compositing.}
Each edited crop is resized to the original dimensions and composited back
into $I$.
For a single-slot record such as $r_1$, the edited region $I'_\text{A}$
replaces the A area in $I$, producing $I'$.
For a two-slot record such as $r_3$, each slot is edited independently,
yielding $I'_\text{C}$ and $I'_\text{D}$, and both are composited into $I$
to produce the final $I'$.

Because prompts are constructed mechanically and each edit is confined to
$s^\text{edit}$, the pipeline is fully automated, reproducible, and
preserves the original visual-search difficulty.

\subsection{Metadata-Grounded QA Construction}
\label{sec:qa-construction}

With $I'$ produced, the final step constructs $Q'$ and $A'$ directly from
$r$.
Because $r$ already encodes the slot, the sampled content, and the
edit-region coordinates, both the question and the answer follow
deterministically.
For an attribute question, the sampled content $c$ is itself the answer, so
there is no gap between what was edited and what the label says.
When constructing $Q'$, the original color is stripped from $t_s$ to keep
the question free of color cues.
For a position question, the answer follows from comparing the spatial
locations of the two selected slots, independent of the edited image
content.
The two constructions are illustrated below using records from
\S\ref{sec:sampling}: $r_1$ for $Q_\text{attr}$ and $r_3$ for
$Q_\text{pos}$ (Figure~\ref{fig:overview}):

\vspace{4pt}
\noindent
\resizebox{\columnwidth}{!}{%
\small
\begin{tabular}{@{}l@{\;\;}p{0.76\columnwidth}@{}}
\toprule
\multicolumn{2}{@{}l}{\textbf{$Q_\text{attr}$}: \;
  $Q'$: ``What color is \{$t_s$\}?'' \; $A' = c$} \\
\midrule
\multicolumn{2}{@{}l}{\textit{Example:} \;
  $r_1 = (\text{A},\, \textit{white})$, \;
  $t_\text{A} = \text{``blue jacket on the cyclist''}$} \\[2pt]
$Q'$ & What color is the jacket on the cyclist? \\
& (A) blue \quad (B) green \quad (C) white \quad (D) yellow \\
$A'$ & (C) white \\
\midrule
\multicolumn{2}{@{}l}{\textbf{$Q_\text{pos}$}: \;
  $Q'$: ``Is \{$c_j$\} to the left or right of \{$c_k$\}?'' \;
  $A'$: $x_j$ vs.\ $x_k$} \\
\midrule
\multicolumn{2}{@{}l}{\textit{Example:} \;
  $r_3 = (\text{C},\, \textit{red bucket},\; \text{D},\, \textit{blue stroller})$, \;
  $x_\text{C} > x_\text{D}$} \\[2pt]
$Q'$ & Is the red bucket to the left or right of the blue stroller? \\
& (A) left \quad (B) right \\
$A'$ & (B) right \\
\bottomrule
\end{tabular}%
}
\vspace{4pt}

\noindent The same $r$ that drives the image edit also fixes the label:
attribute answers come from the sampled content $c$, and position answers
from the annotated slot locations.
Together with human-curated slots (\S\ref{sec:annotation}) and localized
editing (\S\ref{sec:generation}), this ensures that every fresh
$(I', Q', A')$ preserves the original visual-search difficulty, carries a different
correct answer, and has traceable ground truth.

\paragraph{Summary.}
\sysname{} converts a static VQA benchmark into a dynamic one through a
single design principle: a shared sampling record $r$ connects human-curated
edit slots to both the image edit and the ground-truth label.
Each seed $\sigma$ draws a different $r$, producing a fresh $(I', Q', A')$
whose answer is correct by construction.
Because the slots are annotated once and reused across seeds, the same
source images support independent evaluation runs with different visual
keys, different correct answers, and consistent difficulty.

%% file: sections/experiments.tex
\section{Experiments}
\label{sec:experiments}

The gap between original and renewed V*Bench scores
(Figure~\ref{fig:contamination}) suggests that a substantial portion of current
performance reflects memorization rather than genuine visual search.
\sysname{} is designed to close this gap, but a dynamic benchmark must also
earn trust: the difficulty should not shift, scores should remain
consistent across seeds, and the human-in-the-loop design should be
justified rather than replaceable.
We organize our evaluation around three research questions:
Does \sysname{} close the contamination gap (RQ1)?
Does it preserve V*Bench's original difficulty (RQ2)?
And is human annotation necessary, or can a VLM serve as its own annotator
(RQ3)?

\subsection{Setup}
\label{sec:setup}

\paragraph{Setup.}
We annotate all 191 V*Bench \cite{wu2024vstar} images with edit slots
(\S\ref{sec:annotation}), yielding 267 replace and 485 add slots, and
generate three benchmark instances with different seeds (each
${\sim}$114 attribute + ${\sim}$77 position items, edited with
GPT-Image-2 \cite{openai2026gptimage}).
We evaluate eight VLMs from eight families
(Table~\ref{tab:models} in Appendix~\ref{app:protocol}) and, for
difficulty calibration (RQ2), LLaVA-1.5-13B
\cite{DBLP:conf/cvpr/LiuLLL24}.
All models are queried under identical conditions matching the official
V*Bench format: original resolution, no system prompt, temperature~0,
\texttt{max\_tokens}~=~8192 (details in
Appendices~\ref{app:construction} and~\ref{app:protocol}).

\subsection{RQ1: Does \sysname{} Close the Contamination Gap?}
\label{sec:rq1}

\paragraph{Setup.}
We generate three independent \sysname{} benchmark instances from the same
191 source images using three different random seeds $\sigma$. Each seed
produces a fresh sampling record $r$ for every item, yielding different
visual keys and correct answers while keeping the source images fixed.
All eight models (Table~\ref{tab:models}) are evaluated on all three seeds
under the same protocol as the contamination audit.

\begin{table*}[t]
\centering
\small
\setlength{\tabcolsep}{4.2pt}
\begin{tabular}{@{}lccccc@{}}
\toprule
\multicolumn{1}{c}{\textbf{Model}}
  & \multicolumn{2}{c}{\textbf{Static anchors}}
  & \multicolumn{3}{c}{\textbf{\sysname{} renewal}} \\
\cmidrule(lr){2-3}\cmidrule(l){4-6}
\textbf{}
  & \textbf{Original V*}
  & \textbf{Relabel V*}
  & \textbf{Mean $\pm$ std}
  & \textbf{Orig. Drop}
  & \textbf{vs.\ Relabel\,$\approx$\!0} \\
\midrule
Qwen3.6-Plus
  & 90.6 & 74.3 & 75.7 $\pm$ 5.4
  & \drop{14.9} & \near{+1.4} \\
GPT-5.5
  & 86.9 & 75.9 & 77.1 $\pm$ 2.1
  & \drop{9.8} & \near{+1.2} \\
Claude Opus 4.7
  & 78.5 & 59.7 & 67.2 $\pm$ 3.9
  & \drop{11.3} & \near{+7.5} \\
Gemini 3.1 Pro
  & 88.0 & 69.6 & 75.0 $\pm$ 3.3
  & \drop{13.0} & \near{+5.4} \\
Seed-2.0-Lite
  & 91.6 & 74.3 & 79.9 $\pm$ 2.6
  & \drop{11.7} & \near{+5.6} \\
Kimi K2.6
  & 81.7 & 63.4 & 72.6 $\pm$ 0.8
  & \drop{9.1} & \near{+9.2} \\
MiMo v2.5
  & 78.0 & 66.5 & 73.6 $\pm$ 4.7
  & \drop{4.4} & \near{+7.1} \\
Llama 4 Maverick
  & 61.3 & 51.8 & 56.9 $\pm$ 0.6
  & \drop{4.4} & \near{+5.1} \\
\midrule
\textit{Average}
  & \textit{82.1} & \textit{66.9} & \textit{72.2}
  & \drop{9.9} & \near{+5.3} \\
\bottomrule
\end{tabular}
\caption{\sysname{} evaluation (RQ1). \textbf{Relabel~V*}: accuracy when
the same images are paired with fresh human-written questions targeting
different visual keys (Appendix~\ref{app:contamination}).
\textcolor{DropRed}{Red~$\downarrow$}: Original-to-\sysname{} drop.
\textcolor{NearTeal}{Teal}: proximity to Relabel (values near zero =
aligned). \textbf{\sysname{}} reports 3-seed mean $\pm$ standard deviation.
Accuracy in \%; differences in percentage points.}
\label{tab:dynamic}
\end{table*}

\paragraph{Results.}
If the gap identified in Figure~\ref{fig:contamination} reflects memorization, renewing
the visual key should close it: \sysname{} scores should align with
Relabel~V*, the accuracy obtained when the same images are paired with
fresh human-written questions.
Table~\ref{tab:dynamic} supports this.

All eight models drop from Original~V* on \sysname{}.
The largest drops appear in models with the highest Original scores
(Qwen3.6-Plus: $-$14.9~pp from 90.6\%; Gemini: $-$13.0~pp from 88.0\%),
consistent with these models benefiting most from memorization.
Models with lower Original scores show smaller drops
(MiMo and Llama: $-$4.4~pp each), suggesting less contamination exposure.

The Relabel column provides a cross-check.
Two models land almost exactly at the Relabel baseline (Qwen: +1.4~pp;
GPT-5.5: +1.2~pp), indicating that \sysname{} recovers the same difficulty
level as fresh human-written questions.
The remaining models score somewhat above Relabel (mean +5.3~pp),
which we attribute to Relabel using a single fixed question set whereas
\sysname{} benefits from sampling diversity.
Cross-seed standard deviations range from 0.6 to 5.4~pp, confirming that
individual seeds produce consistent measurements
(Figure~\ref{fig:dynamic_breakdown} in Appendix~\ref{app:analysis}).

The alignment with Relabel is consistent with the drop reflecting
contamination rather than a difficulty shift.
A per-category decomposition (Figure~\ref{fig:dynamic_breakdown}) reinforces
this interpretation: the drop
concentrates in attribute questions, where memorizing a single visual
property provides the most direct shortcut.

\paragraph{Chain-of-thought amplifies contamination.}
A natural concern is whether chain-of-thought (CoT) prompting could
circumvent \sysname{}'s decontamination: if step-by-step reasoning helps
a model \emph{genuinely} analyze the edited image, CoT might recover
performance on \sysname{} just as it boosts scores on static benchmarks.
To test this, we select three reasoning-capable models, turn off their
internal thinking mode so that all reasoning must appear in the visible
output, and compare two prompt variants: the standard direct-answer
suffix (\emph{Direct}) and a step-by-step instruction (\emph{CoT}:
``Think step by step, then answer\ldots'')
(Figure~\ref{fig:capability}a; setup in Appendix~\ref{app:protocol},
full results in Table~\ref{tab:cot-detail}).

On static benchmarks, CoT inflates scores substantially:
Original~V* rises by up to 6.8~pp, and, critically, Relabel~V* rises by
an even larger margin (up to 10.5~pp).
The fact that Relabel also benefits is revealing: Relabel uses
\emph{fresh} questions on the same images, so the boost cannot come from
memorizing the original question--answer pair. Instead, CoT appears to
help models retrieve benchmark-associated knowledge more broadly,
recognizing familiar images and leveraging any static association between
image content and answer patterns.

On \sysname{}, this mechanism breaks down. The CoT effect turns negative
for all three models (approximately $-$0.5 to $-$3.1~pp), meaning step-by-step
reasoning actively hurts on renewed visual keys.
The contrast is clear: CoT boosts most static evaluations yet provides
no benefit once the visual key is renewed.
This suggests that CoT's apparent strength on V*Bench is not visual
reasoning but structured recall of benchmark-associated knowledge.
By stripping away this shortcut, \sysname{} reveals that explicit
step-by-step reasoning does not aid genuine visual search: the task is
fundamentally perceptual, and CoT's contribution was an artifact of
contamination.

\begin{figure}[t]
  \centering
  \includegraphics[width=\columnwidth]{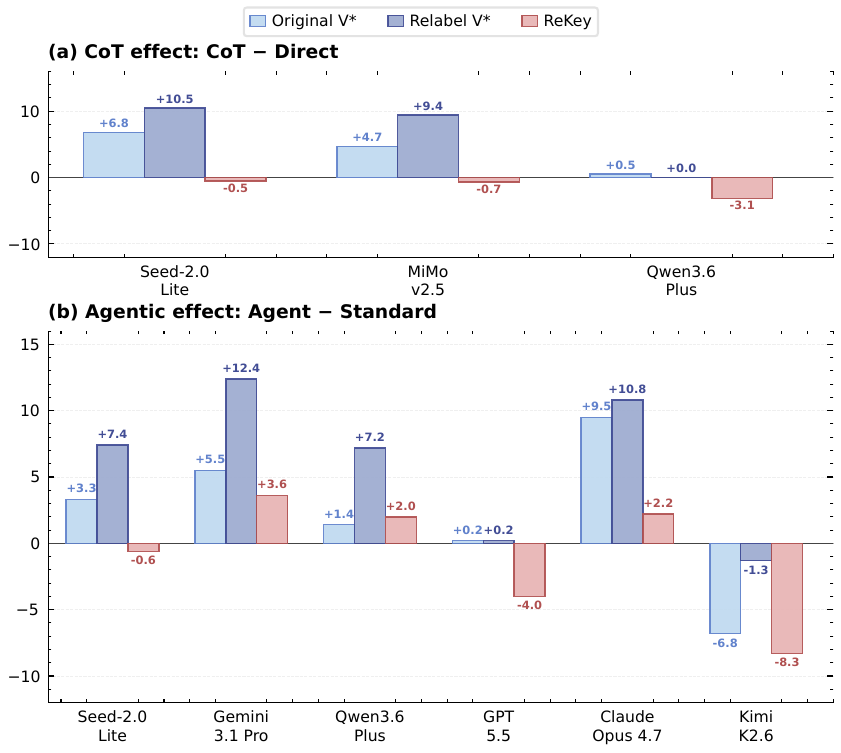}
  \caption{Two capability boosts that disappear under \sysname{} (RQ1).
  \textbf{(a)}~CoT effect (CoT~$-$~Direct): three reasoning models
  with internal thinking disabled. \textbf{(b)}~Agentic effect
  (Agent~$-$~Standard): six models in a ReAct harness with
  iterative zoom. Both capabilities boost most static evaluations (Original
  and Relabel, blue bars) but the effect shrinks or turns negative on
  \sysname{} (red). Effects are in percentage points.}
  \label{fig:capability}
\end{figure}

\paragraph{Agentic VLMs.}
A parallel concern is whether iterative tool use could circumvent
\sysname{}'s decontamination: if a zoom tool helps a model
systematically search the image, it might locate renewed visual keys
just as effectively as static ones.
To test this, we evaluate six models in a ReAct harness that provides a
crop-and-zoom tool for iterative inspection
(Figure~\ref{fig:capability}b; full results in
Table~\ref{tab:agentic-detail}).

On static benchmarks, tool use raises accuracy substantially
(Original +1.4--9.5\,pp, Relabel +7.2--12.4\,pp for four of six
models).
Since Relabel uses fresh questions on the same images, the tool
provides genuine visual-search capability beyond memorized answers,
guided by the model's familiarity with the scene.

On \sysname{}, this advantage does not transfer: the tool boost shrinks
to $-$4.0 to $+$3.6\,pp for five of six models.
Both Relabel and \sysname{} require locating a different visual key,
but Relabel preserves the original image while
\sysname{} locally edits it.
On Relabel, the model's scene understanding guides the tool to
relevant regions; on \sysname{}, the visual key itself is novel, so
even zooming into the correct area does not yield a recognizable
target.
This shows that \sysname{} is robust against tool-augmented evaluation.

\subsection{RQ2: Does \sysname{} Preserve the Original Difficulty?}
\label{sec:rq2}

The accuracy drop from Original~V* to \sysname{} could reflect a difficulty
shift rather than contamination removal. We examine this from three angles.

\paragraph{Edit-region scale.}
By design, each edit modifies only the slot region $s_k^\text{edit}$, whose
median area is approximately 3,500~px$^2$ in a
${\sim}$2250$\times$1500 image, roughly 0.1\% of the total pixels.
As described in \S\ref{sec:annotation}, replace slots target the same
objects that V*Bench's original questions ask about, so their edit regions
directly reflect V*Bench's small-target scale.
Human-annotated add regions are of similar size
(${\sim}$3,300~px$^2$ vs.\ ${\sim}$3,800~px$^2$ for replace;
Table~\ref{tab:edit-size}), confirming that annotators select insertion
locations that match the original difficulty standard.
The vast majority of the image remains untouched, limiting the scope for
difficulty shift.

\paragraph{Model calibration.}
To verify empirically, we evaluate LLaVA-1.5-13B, the strongest
open-source baseline in the original V*Bench paper \cite{wu2024vstar}.
This model anchors V*Bench's published difficulty and carries lower
contamination risk, so a large drop would signal a task change rather than
decontamination.
Table~\ref{tab:difficulty} shows that LLaVA-1.5-13B drops by only 6.1~pp
overall, smaller than most frontier models in RQ1 (9.1--14.9~pp for six
of eight models).
Position accuracy is nearly unchanged ($-$0.4~pp), which is particularly
important because spatial relations are the aspect most sensitive to
editing artifacts.
The attr/pos asymmetry
(Figure~\ref{fig:dynamic_breakdown} in Appendix~\ref{app:analysis})
reinforces this: attribute questions drop more because memorizing a
single color provides a direct shortcut, while position questions
barely change.
Together, the small edit-region footprint, the modest LLaVA calibration
drop, and the stability of position accuracy indicate that the RQ1
drops are contamination-driven rather than difficulty-driven.

\begin{table}[t]
\centering
\small
\setlength{\tabcolsep}{4.6pt}
\begin{tabular}{@{}lccc@{}}
\toprule
\textbf{Split} & \textbf{Original V*} & \textbf{\sysname{}}
& \textbf{Drop} \\
\midrule
Overall & 48.68 & 42.6 $\pm$ 2.4 & \drop{6.1} \\
Attr & 43.47 & 33.2 $\pm$ 3.9 & \drop{10.3} \\
Pos & 56.57 & 56.2 $\pm$ 3.3 & \drop{0.4} \\
\bottomrule
\end{tabular}
\caption{Difficulty calibration on LLaVA-1.5-13B (RQ2). \textbf{Original V*}
uses the published result from \citet{wu2024vstar}; \textbf{\sysname{}}
reports the 3-seed mean $\pm$ standard deviation. Accuracy in \%; drops in
percentage points.}
\label{tab:difficulty}
\end{table}

\subsection{RQ3: Is Human Annotation Necessary?}
\label{sec:rq3}

\paragraph{Setup.}
A natural question is whether the human annotation stage can be
replaced by a VLM, reducing cost and increasing scale.
Vision-Language Bootstrapping \cite{yang2025vlb}, for instance, uses
GPT-4V to select regions for perturbation.
We prompt GPT-5.5 with each V*Bench image and the same annotation
instructions given to human annotators.
However, only 33.6\% of the VLM's replace annotations are correctly
localized (Table~\ref{tab:edit-size}): GPT-5.5 frequently places
bounding boxes at incorrect locations, failing to locate the small,
peripheral targets that V*Bench requires.
We therefore construct the VLM-annotated benchmark from add slots only
(189 items) and evaluate all eight models on both benchmarks.

\begin{figure}[t]
  \centering
  \includegraphics[width=0.85\columnwidth]{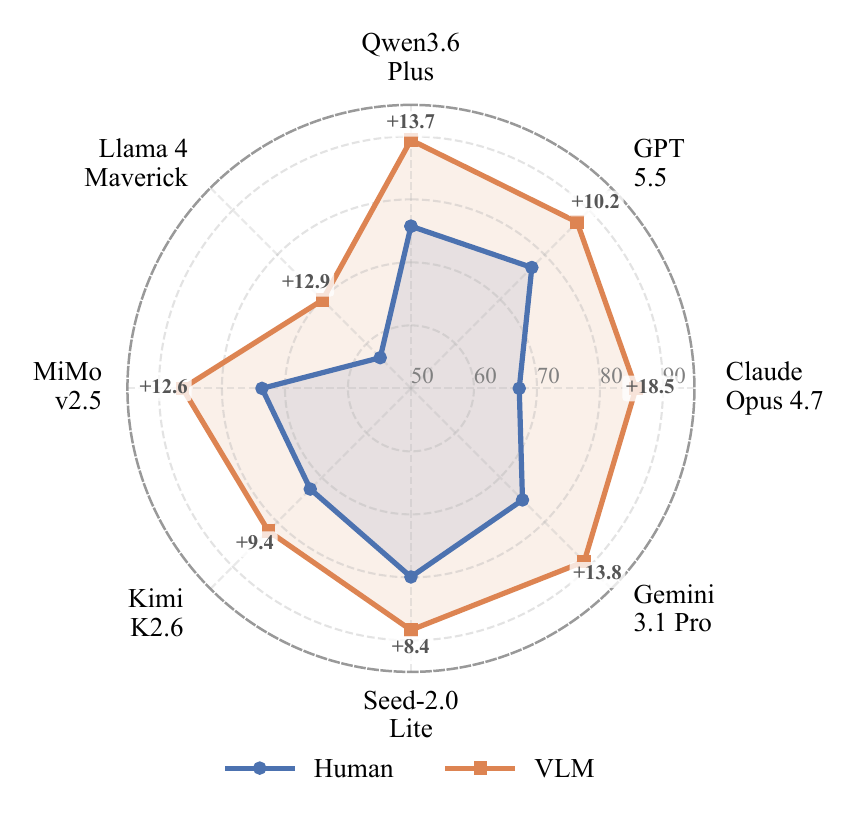}
  \caption{Human vs.\ VLM annotation (RQ3). \textbf{Human} (blue): mean
  accuracy on the human-annotated \sysname{} benchmark (3 seeds).
  \textbf{VLM} (orange): accuracy on the GPT-5.5-annotated benchmark
  using add slots only. All eight models score +8.4--18.5~pp higher on
  the VLM-annotated version. Accuracy in \%.}
  \label{fig:vlm-annotator}
\end{figure}

\paragraph{Results.}
All eight models score 8.4--18.5~pp higher on the VLM-annotated
benchmark (Figure~\ref{fig:vlm-annotator}).
Table~\ref{tab:edit-size} reveals the cause: human-annotated add
regions (${\sim}$3,300~px$^2$) match the small-target scale of V*Bench's
replace regions (${\sim}$3,800~px$^2$), but VLM-annotated add regions are
3.2$\times$ larger (${\sim}$10,500~px$^2$), because GPT-5.5 favors
prominent, spacious areas where inserted objects are easy to spot.
This saliency bias systematically inflates accuracy, undermining the
difficulty preservation that human curation provides.

\begin{table}[t]
\centering
\small
\begin{tabular}{@{}lcc@{}}
\toprule
& \textbf{Replace} & \textbf{Add} \\
\midrule
Human size (px$^2$) & 3{,}843 & 3{,}256 \\
VLM size (px$^2$) & 3{,}531 & 10{,}504 \\
Size ratio & 0.92$\times$ & \textbf{3.2$\times$} \\
Usability & \textcolor{DropRed}{33.6\%} & \textcolor{NearTeal}{${\sim}$94\%} \\
\bottomrule
\end{tabular}
\caption{VLM annotation quality by slot type (RQ3). Human replace regions
serve as a difficulty baseline: they target the original V*Bench visual
keys and reflect the benchmark's small-target scale. \textbf{Size ratio}:
VLM median / Human median. \textbf{Usability}: fraction of VLM-annotated
slots whose bounding box correctly covers the intended target (manual
inspection). Replace annotations are mostly unusable
(\textcolor{DropRed}{33.6\%}); add annotations are largely usable
(\textcolor{NearTeal}{${\sim}$94\%}) but 3.2$\times$ larger than the
human baseline.}
\label{tab:edit-size}
\end{table}

%% file: sections/conclusion.tex
\section{Conclusion}
\label{sec:conclusion}

V*Bench scores are substantially higher than warranted, consistent
with contamination.
\sysname{} addresses this by treating the visual key as the renewable
unit: human annotators mark edit slots once, and a rule-based pipeline
generates fresh instances whose ground truth is correct by
construction.
Our experiments show that \sysname{} closes the contamination gap while
preserving difficulty, and that neither chain-of-thought prompting nor
agentic tool use restores the static-benchmark gains in our experiments.
A VLM self-annotation ablation further supports the need for human
curation, as automated annotation introduces systematic saliency bias.
More broadly, one-time human curation combined with automated renewal
shows that curated visual benchmarks need not be disposable: with the
right design, the same source images can support fresh, reliable
evaluation as models and training data continue to evolve.

%% file: sections/limitations.tex
\section*{Limitations}

\paragraph{Single generator.}
All edits use GPT-Image-2. Instruction-based editing does not guarantee
pixel preservation outside the edit region: visible seams at crop
boundaries and occasional object size mismatches are observed. Switching
to a different editing backend may introduce new artifact patterns that
require re-validation.

\paragraph{Attribute coverage.}
The current implementation varies only color. The method framework
supports other visual attributes (material, texture), but these require
corresponding attribute pools and have not been validated.

\paragraph{Cost.}
Annotating edit slots for all 191 V*Bench images required approximately
16 person-hours (two annotators, one day each).
This is a one-time effort: once annotated, the same slots support
many independent benchmark instances across different seeds.
Generating one instance costs approximately \$21 in GPT-Image-2 API
calls, and evaluating eight models on one seed costs approximately
\$14 in VLM API calls.

\paragraph{Difficulty control.}
The pipeline's design implicitly preserves visual-search difficulty:
replace slots keep the target at its original location, and add slots
use human-curated regions that match V*Bench's small-target scale
(RQ1 and RQ2 confirm this at the aggregate level).
However, the pipeline treats all items uniformly and provides no
mechanism for explicitly controlling or grading item difficulty.
It cannot generate a harder or easier variant on demand, nor
distinguish easy items from hard ones before evaluation.
This limits applications that require fine-grained difficulty
profiling, such as adaptive testing or curriculum-based evaluation.
Introducing difficulty-aware sampling, for instance by conditioning on
slot location, object size, or scene clutter, is a natural direction
for future work.

\paragraph{V*Bench scope.}
We validate \sysname{} exclusively on V*Bench (191 images). Generalization
to other visual benchmarks with different image characteristics or question
formats has not been tested.

%% file: appendices/appendix.tex
\section{V*Bench Contamination Audit}
\label{app:contamination}

\begin{figure*}[t]
  \centering
  \includegraphics[width=0.95\textwidth]{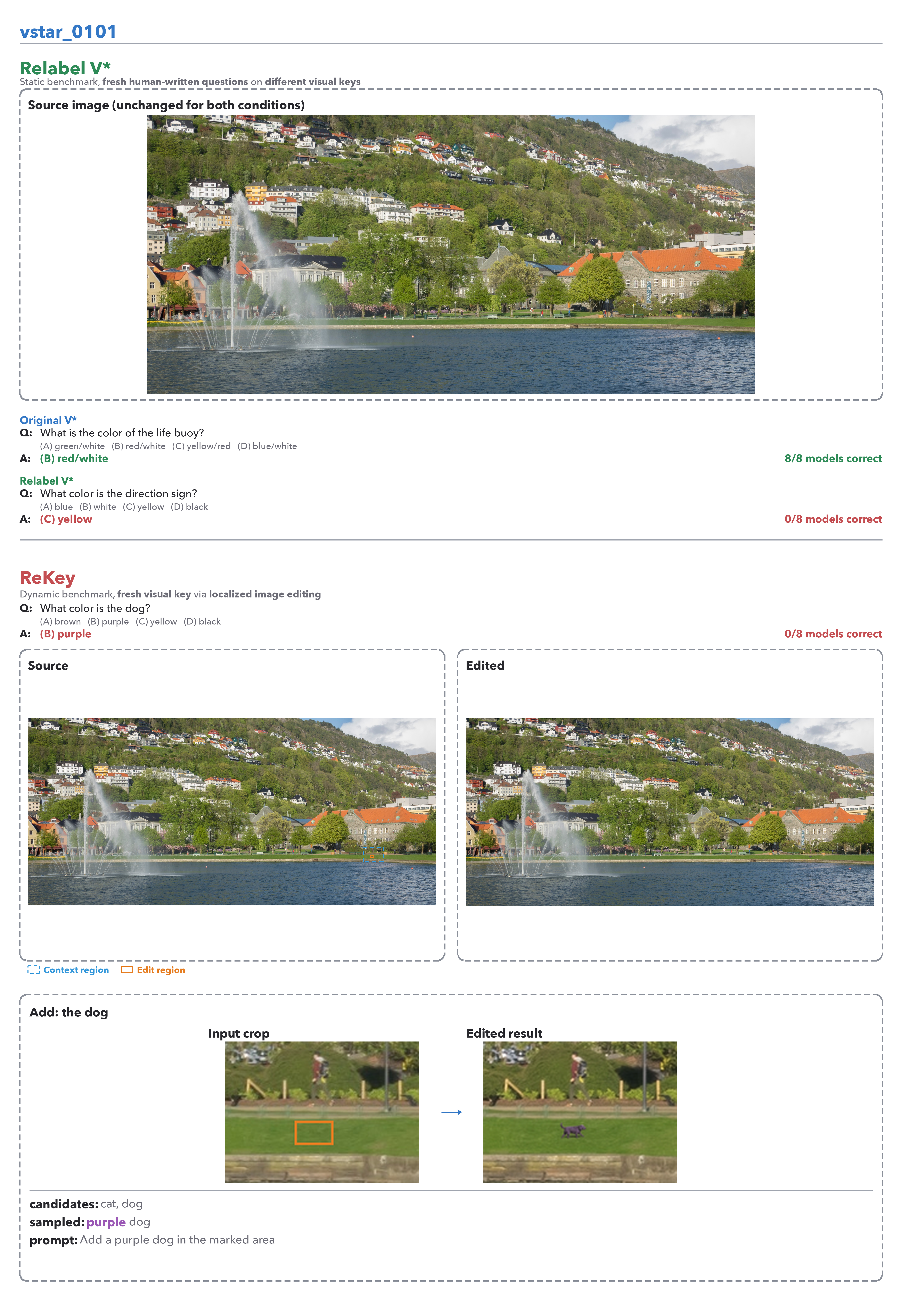}
  \caption{Two ways to break memorization on the same V*Bench image
  (vstar\_0101). \textbf{Relabel~V*} (top) keeps the source image
  unchanged and pairs it with a fresh human-written question about a
  different visual key (direction sign instead of life buoy).
  \textbf{\sysname{}} (bottom) inserts a new visual key (purple dog) via
  localized editing and derives both the question and the answer from
  the sampling record~$r$. All eight models answer the original question
  correctly (8/8) but fail on both renewed conditions (0/8).}
  \label{fig:relabel-vs-rekey}
\end{figure*}

To measure how much current V*Bench performance reflects memorization,
we construct \emph{Relabel~V*}: the same 191 images paired with fresh
human-written questions that target different visual keys while
preserving the original difficulty regime (small, peripheral targets in
cluttered scenes).

Figure~\ref{fig:relabel-vs-rekey} walks through a concrete example.
On vstar\_0101, all eight models correctly answer the original question
(``What color is the life buoy?''\ $\rightarrow$ red/white, 8/8).
The life buoy is a small, peripheral detail in a cluttered lakeside
scene, exactly the kind of target V*Bench is designed to test.
Yet when a human annotator writes a fresh question about a different
visual key in the same image (``What color is the direction sign?''\
$\rightarrow$ yellow), every model fails (0/8).
The image has not changed; only the question target has.
This sharp drop on a single item suggests that models recalled a
memorized answer rather than locating the visual key through genuine
visual search.
The bottom half of the figure shows the \sysname{} counterpart: a
purple dog is inserted via localized editing, and the question and
answer are derived from the sampling record.
All models fail again (0/8), but unlike Relabel, \sysname{} produces
this effect automatically from a single annotation.

The pattern illustrated by vstar\_0101 is not isolated.
Figure~\ref{fig:contamination} aggregates the results across all 191
items and eight models: every model's overall accuracy drops when the
same images are paired with fresh questions targeting different visual
keys.
Seven of eight drops are statistically significant ($p < 0.01$,
McNemar's exact test); only Llama~4~Maverick ($-$9.5\,pp, $p = 0.09$)
falls short of significance, consistent with its lower baseline and
lower contamination exposure.
The largest drops appear in models with the highest Original scores
(Claude~Opus~4.7: $-$18.8\,pp from 78.5\%;
Gemini~3.1~Pro ($-$18.4\,pp) and Kimi~K2.6 ($-$18.3\,pp)),
consistent with these models benefiting most from memorization.
A detailed breakdown by question type is provided in
Appendix~\ref{app:analysis}.

\section{Annotation and Construction Details}
\label{app:construction}

\paragraph{Slot design rationale.}
The two slot modes serve complementary roles in breaking memorization
while preserving the visual-search task.

Replace slots target the object that the original V*Bench question asks
about.
Most V*Bench questions reference a unique object in the scene (e.g.\
``What color is the life buoy?''), so modifying that specific object
directly invalidates the memorized answer while largely preserving the
search task: the model must still locate the same small, peripheral
target in the same cluttered scene.
Because the object's identity and location are unchanged, a
performance drop is consistent with loss of the memorized answer
rather than a change in difficulty.

Add slots introduce a new object that is absent from the source image,
reducing the chance of a fixed scene-object association in training
data.
Annotators curate the candidate set $\mathcal{C}_k$ to contain only
objects that are plausible in the local context, absent from the source
image, and small enough to preserve the visual-search difficulty
(\S\ref{sec:annotation}).
The absence constraint is enforced during sampling: each candidate is
checked against all replace targets in the scene (both the full
description and its last word) and excluded if a match is found.
For position questions, an additional spatial-separation constraint
requires the two selected slot centers to be at least 20\,px apart,
ensuring that the directional answer (left/right or above/below) is
unambiguous.
Together, these constraints ensure that answering a \sysname{} question
requires genuine visual search rather than recalling typical scene
compositions.

\paragraph{Position-question format.}
Position questions ask about the spatial relation between two objects,
so both objects must be unambiguously identifiable in the question text.
The format follows from the slot design above.

A replace object retains its identity and location; only its color
changes.
Because the scene may contain other instances of the same category, the
question includes the \emph{new} color to let the model distinguish the
edited object from its surroundings
(e.g.\ ``the \textbf{orange} phone'' after recoloring from black).
An add object, by contrast, is the only instance of its category in the
scene (guaranteed by the absence constraint), so the category name alone
is sufficient (e.g.\ ``the cat'').
Including the color for add objects would be redundant and could
introduce an unintended cue that makes the object easier to locate.

\section{Evaluation Protocol Details}
\label{app:protocol}

\begin{table}[t]
\centering
\small
\setlength{\tabcolsep}{3pt}
\resizebox{\columnwidth}{!}{%
\begin{tabular}{@{}lcl@{}}
\toprule
\textbf{Model} & \textbf{Type} & \textbf{Reasoning} \\
\midrule
\modelname{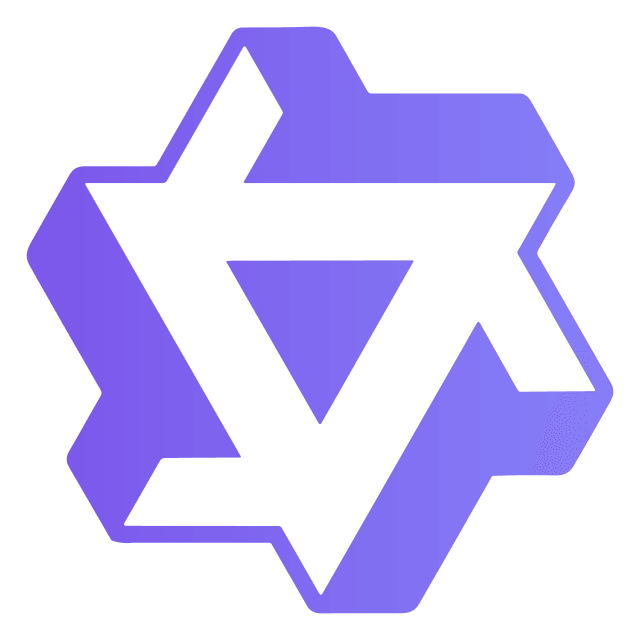}{Qwen3.6-Plus~\citep{qwen3.6plus}}
    & \XSolidBrush & Always-on CoT \\
\modelname{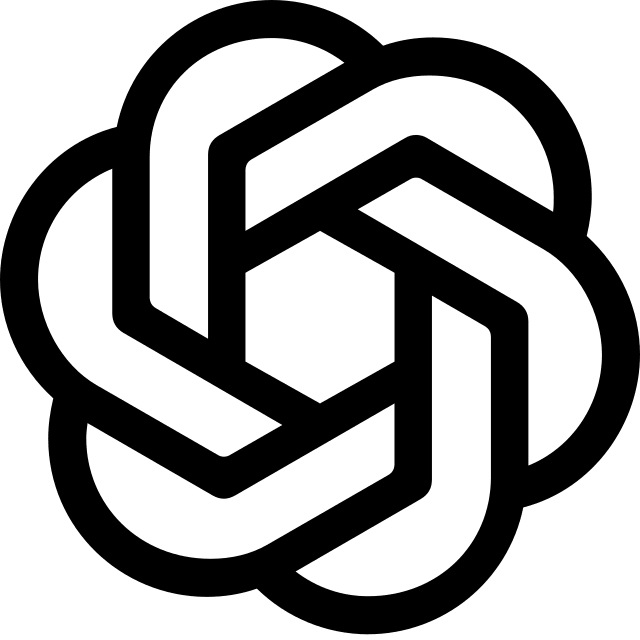}{GPT-5.5~\citep{gpt5.5}}
    & \XSolidBrush & Medium \\
\modelname{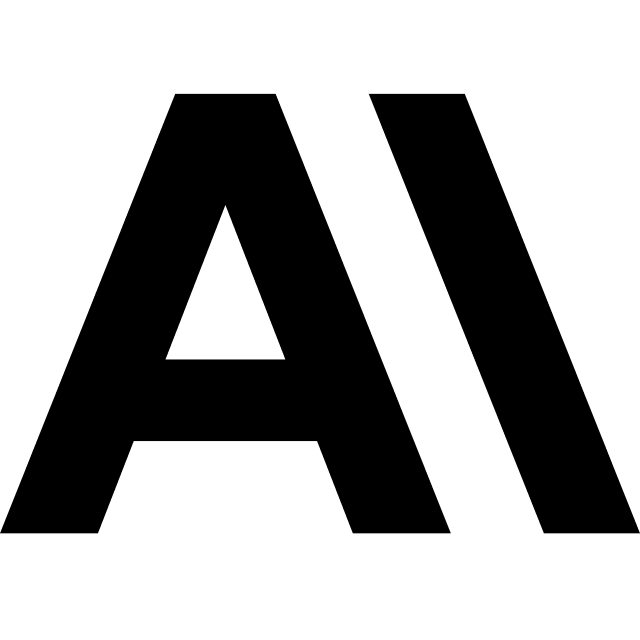}{Claude Opus 4.7~\citep{claudeopus4.7}}
    & \XSolidBrush & Adaptive \\
\modelname{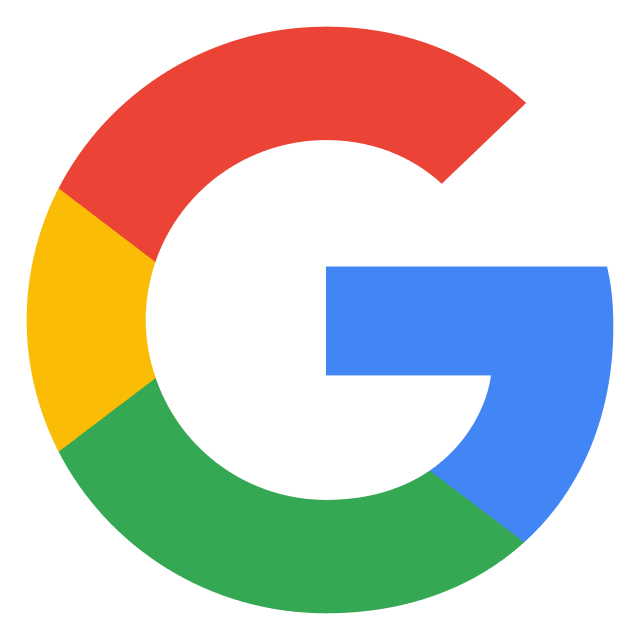}{Gemini 3.1 Pro~\citep{gemini3.1pro}}
    & \XSolidBrush & Medium \\
\modelname{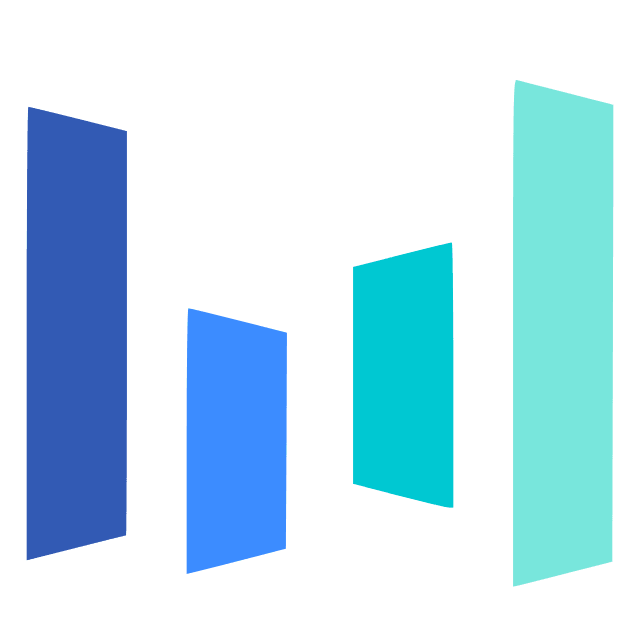}{Seed-2.0-Lite~\citep{seed2.0lite}}
    & \XSolidBrush & None \\
\modelname{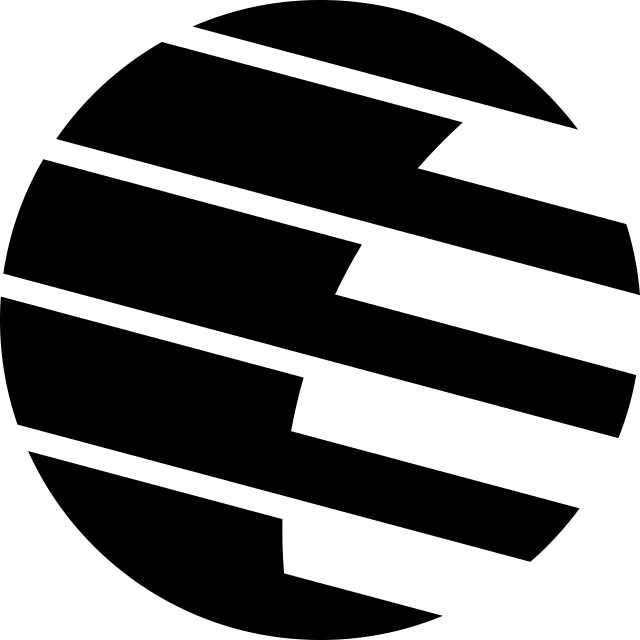}{Kimi K2.6~\citep{kimik2.6}}
    & \XSolidBrush & Default \\
\modelname{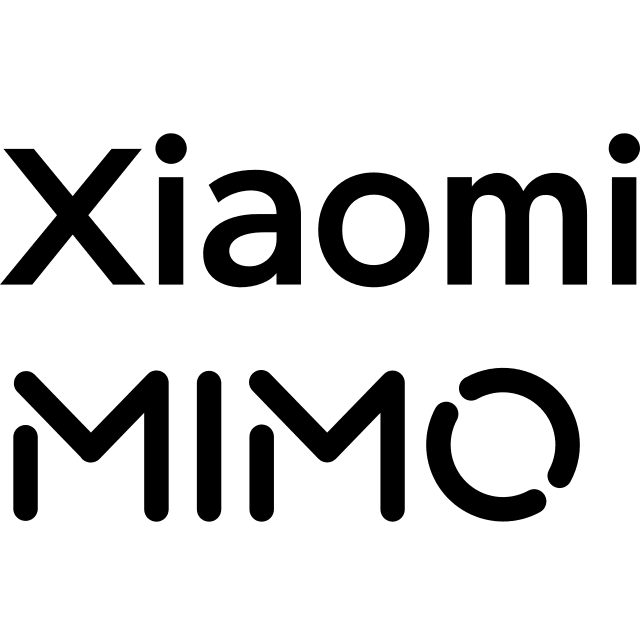}{MiMo v2.5~\citep{xiaomi2026mimov25}}
    & \Checkmark & Default \\
\modelname{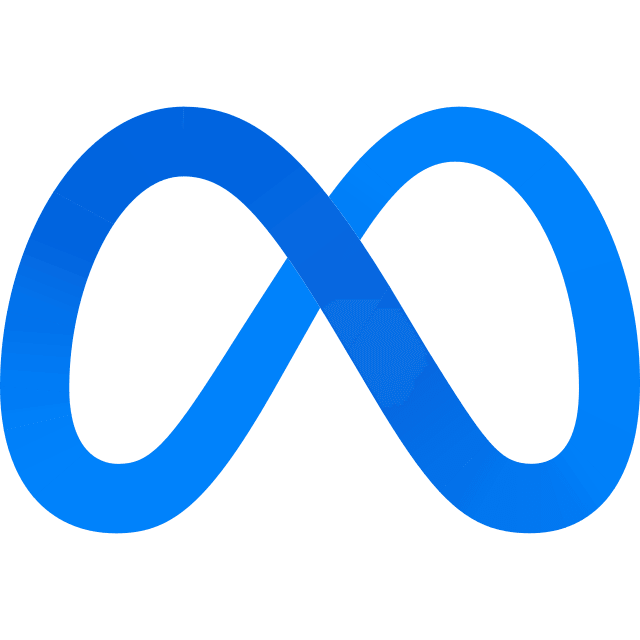}{Llama 4 Maverick~\citep{llama4maverick}}
    & \Checkmark & None \\
\bottomrule
\end{tabular}%
}
\caption{Evaluated models spanning eight families. \textbf{Type}:
\XSolidBrush~closed-weight, \Checkmark~open-weight.
\textbf{Reasoning}: provider-default strategy (not overridden).}
\label{tab:models}
\end{table}

All models are queried through the OpenRouter API under identical
conditions: original image resolution, no system prompt, temperature~0,
and \texttt{max\_tokens}~=~8192.
Each provider exposes a default reasoning strategy that we do not
override (Table~\ref{tab:models}); this mirrors a realistic deployment
setting where the user accepts the provider's default.
Source images are re-encoded as JPEG (quality~95) for API
compatibility; no resizing is applied.

\paragraph{Prompt format.}
Each item is presented as a single-turn message containing the image
and a text prompt.
For the standard (direct) evaluation, the prompt follows the official
V*Bench format:

\vspace{2pt}
\noindent
\small
\texttt{\{question\} (A) \{a\} (B) \{b\} \ldots\ Answer with the
option's letter from the given choices directly.}
\normalsize
\vspace{2pt}

\noindent For the CoT experiments (\S\ref{sec:rq1}), the suffix is
replaced with:

\vspace{2pt}
\noindent
\small
\texttt{Think step by step, then answer with the option's letter from
the given choices.}
\normalsize
\vspace{2pt}

\noindent Responses are parsed by extracting the \emph{last} option
letter (A/B/C/D) in the output, which avoids matching letters that
appear in intermediate reasoning text.

\paragraph{CoT experimental setup.}
The chain-of-thought analysis in \S\ref{sec:rq1} uses three
reasoning-capable models with their internal thinking mode disabled
(OpenRouter \texttt{reasoning.effort}~=~\texttt{none}), so that all
reasoning must appear in the visible output.
Two prompt variants are compared: the standard direct-answer suffix
above (\emph{Direct}) and a step-by-step instruction (\emph{CoT})
shown above.
This setup isolates the effect of explicit chain-of-thought reasoning
from provider-internal thinking tokens.

\section{CoT and Agentic Detailed Results}
\label{app:cot-agentic}

Table~\ref{tab:cot-detail} reports the CoT experiment from
\S\ref{sec:rq1}: three reasoning models with internal thinking
disabled, evaluated under Direct and CoT prompting across three
benchmark conditions.
Table~\ref{tab:agentic-detail} reports the agentic experiment: six
models in a ReAct harness with iterative zoom, compared with their
standard (no-tool) counterparts.

\begin{table}[ht]
\centering
\footnotesize
\setlength{\tabcolsep}{3pt}
\begin{tabular}{@{}llccc@{}}
\toprule
\textbf{Model} & \textbf{Mode}
  & \textbf{Orig.} & \textbf{Relab.} & \textbf{\sysname{}} \\
\midrule
Seed-2.0-Lite & Direct & 82.7 & 62.8 & 79.8 $\pm$ 3.0 \\
              & CoT    & 89.5 & 73.3 & 79.2 $\pm$ 4.3 \\
\midrule
MiMo v2.5    & Direct & 74.3 & 55.0 & 68.2 $\pm$ 4.7 \\
              & CoT    & 79.1 & 64.4 & 67.5 $\pm$ 3.1 \\
\midrule
Qwen3.6-Plus & Direct & 85.9 & 66.5 & 78.9 $\pm$ 4.1 \\
              & CoT    & 86.4 & 66.5 & 75.7 $\pm$ 4.5 \\
\bottomrule
\end{tabular}
\caption{CoT vs.\ Direct accuracy with internal thinking disabled
(\S\ref{sec:rq1}). \textbf{Orig.}: Original~V*. \textbf{Relab.}:
Relabel~V*. \textbf{\sysname{}}: 3-seed mean $\pm$ standard deviation.
Accuracy in~\%.}
\label{tab:cot-detail}
\end{table}

\begin{table}[ht]
\centering
\footnotesize
\setlength{\tabcolsep}{2.8pt}
\begin{tabular}{@{}lcccccc@{}}
\toprule
& \multicolumn{2}{c}{\textbf{Original}}
& \multicolumn{2}{c}{\textbf{Relabel}}
& \multicolumn{2}{c}{\textbf{\sysname{}}} \\
\cmidrule(lr){2-3}\cmidrule(lr){4-5}\cmidrule(l){6-7}
\textbf{Model}
  & \textbf{Ag.}
  & \textbf{Base}
  & \textbf{Ag.}
  & \textbf{Base}
  & \textbf{Ag.}
  & \textbf{Base} \\
\midrule
Seed-2.0-Lite    & 94.9 & 91.6 & 81.7 & 74.3 & 79.3\tiny$\pm$3.4 & 79.9\tiny$\pm$2.6 \\
Gemini 3.1 Pro   & 93.5 & 88.0 & 82.0 & 69.6 & 78.6\tiny$\pm$2.0 & 75.0\tiny$\pm$3.3 \\
Qwen3.6-Plus     & 92.0 & 90.6 & 81.5 & 74.3 & 77.7\tiny$\pm$4.0 & 75.7\tiny$\pm$5.4 \\
GPT-5.5          & 87.1 & 86.9 & 76.1 & 75.9 & 73.1\tiny$\pm$2.0 & 77.1\tiny$\pm$2.1 \\
Claude Opus 4.7  & 88.0 & 78.5 & 70.5 & 59.7 & 69.4\tiny$\pm$2.6 & 67.2\tiny$\pm$3.9 \\
Kimi K2.6        & 74.9 & 81.7 & 62.1 & 63.4 & 64.3\tiny$\pm$2.0 & 72.6\tiny$\pm$0.8 \\
\bottomrule
\end{tabular}
\caption{Agentic vs.\ standard VLMs (\S\ref{sec:rq1}).
\textbf{Ag.}: ReAct harness with iterative zoom tool.
\textbf{Base}: standard single-pass evaluation
(Table~\ref{tab:dynamic}). \sysname{} columns report 3-seed mean $\pm$ standard deviation;
Original and Relabel are from single evaluations.
Accuracy in~\%.}
\label{tab:agentic-detail}
\end{table}

\section{Additional Analysis}
\label{app:analysis}

\begin{figure}[H]
  \centering
  \includegraphics[width=0.75\columnwidth]{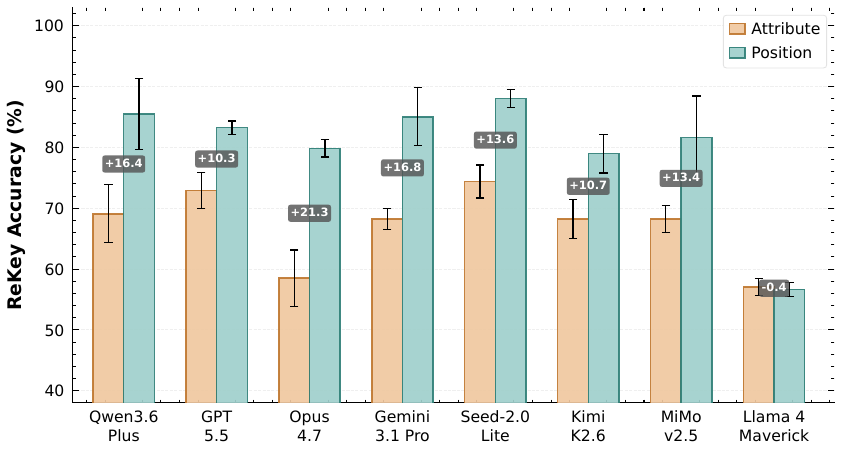}
  \caption{\sysname{} accuracy by question type (3-seed mean $\pm$
  $\sigma$). Gray tags show the position--attribute gap in percentage points.
  Llama~4~Maverick is the only model without a positive gap, consistent
  with its low contamination exposure. Accuracy in~\%.}
  \label{fig:dynamic_breakdown}
\end{figure}

\paragraph{Attribute vs.\ position asymmetry.}
Figure~\ref{fig:dynamic_breakdown} decomposes \sysname{} accuracy into
attribute ($Q_\text{attr}$) and position ($Q_\text{pos}$) questions.
Compared to Original~V* (Table~\ref{tab:dynamic}), attribute accuracy
drops 14.8~pp on average while position drops only 2.6~pp, a
5.7$\times$ asymmetry discussed in \S\ref{sec:rq1}.
Three factors explain why attribute questions are more susceptible.

First, attribute answers are compact, single-token facts (e.g.\
``blue'') that are easy to memorize from a fixed benchmark.
\sysname{} replaces each memorized value with a freshly sampled color,
so the cached answer has only a $1/|\mathcal{A}|$ chance
(${\approx}$9\%) of remaining correct.
Position answers, by contrast, depend on the spatial relation between
two objects---a relational judgment that is harder to encode as a
training-data shortcut.

Second, the two question types differ in baseline chance.
Attribute questions offer four options (25\% chance), leaving a wide
margin for contamination-driven inflation.
Position questions offer two (50\%), compressing the range over which
memorization can inflate scores.

Third, \sysname{}'s renewal mechanism preserves spatial layout by
construction.
Replace slots keep the target object at its original location;
add slots insert objects at annotated coordinates.
The relative position of any slot pair is therefore fixed across seeds,
so position accuracy reflects genuine spatial reasoning rather than
memorized layout.
The small mean position drop ($-$2.6~pp) provides independent support
for the difficulty-preservation argument (RQ2): spatial relations are
the aspect most sensitive to editing artifacts, and their stability
indicates that the localized editing process preserves the scene's
spatial structure.

Llama~4~Maverick is the sole exception to the pattern: its
position--attribute gap is $-$0.4~pp, compared to $+$10.3 to $+$21.3
for all other models
(Figure~\ref{fig:dynamic_breakdown}).
This is consistent with Llama having the smallest and only
non-significant contamination drop ($-$9.5~pp, $p = 0.09$;
Figure~\ref{fig:contamination}): without a strong memorization advantage
on attribute questions, its attr and pos accuracy are roughly equal
under \sysname{}.

\section{Qualitative Examples}
\label{app:qualitative}

We examine individual \sysname{} items to illustrate how visual-key
renewal breaks memorization while preserving visual-search difficulty.
The two slot modes provide complementary mechanisms: inserting a novel
object (add mode) and recoloring an existing target (replace mode).

\begin{figure*}[t]
  \centering
  \includegraphics[width=\textwidth]{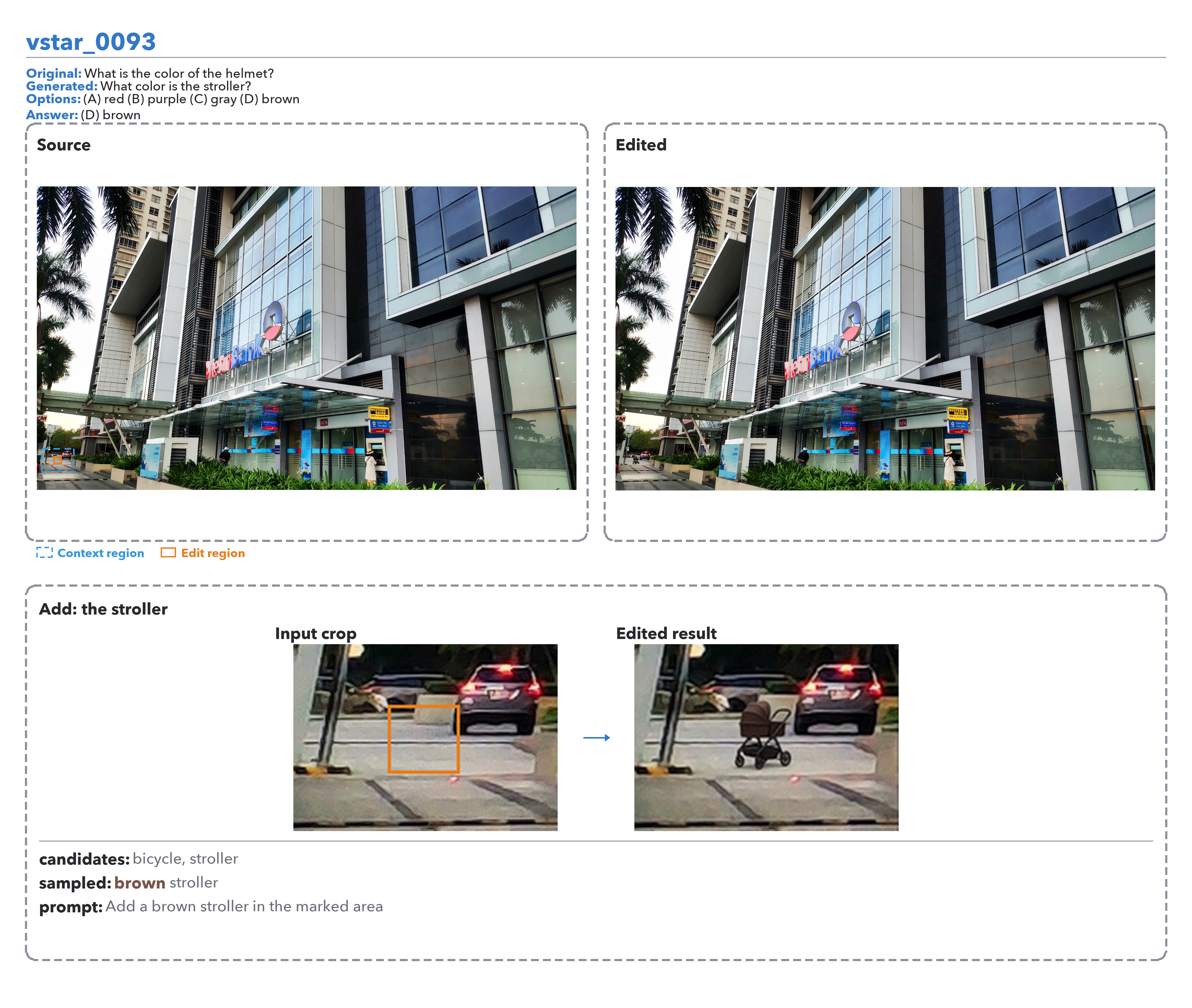}
  \caption{Success: add mode (vstar\_0093, bench2). A small brown
  stroller is inserted in a parking-area corner. Original accuracy:
  8/8; \sysname{} accuracy: 0/8.}
  \label{fig:succ-add}
\end{figure*}

\paragraph{Add mode: novel object in a cluttered scene.}
Figure~\ref{fig:succ-add} shows vstar\_0093.
The original question asks about a helmet's color in an urban street
scene, and all eight models answer correctly (8/8).
\sysname{} inserts a small brown stroller in the corner of a parking
area; the generated question asks its color.
No model answers correctly (0/8).
The stroller is small, placed at the periphery between a parked car
and a crosswalk, and surrounded by visual clutter (signage, building
facades, street furniture).
Because the object is absent from the source image, no model has a
prior association between this scene and the stroller, so answering
requires genuine visual search.

\begin{figure*}[t]
  \centering
  \includegraphics[width=\textwidth]{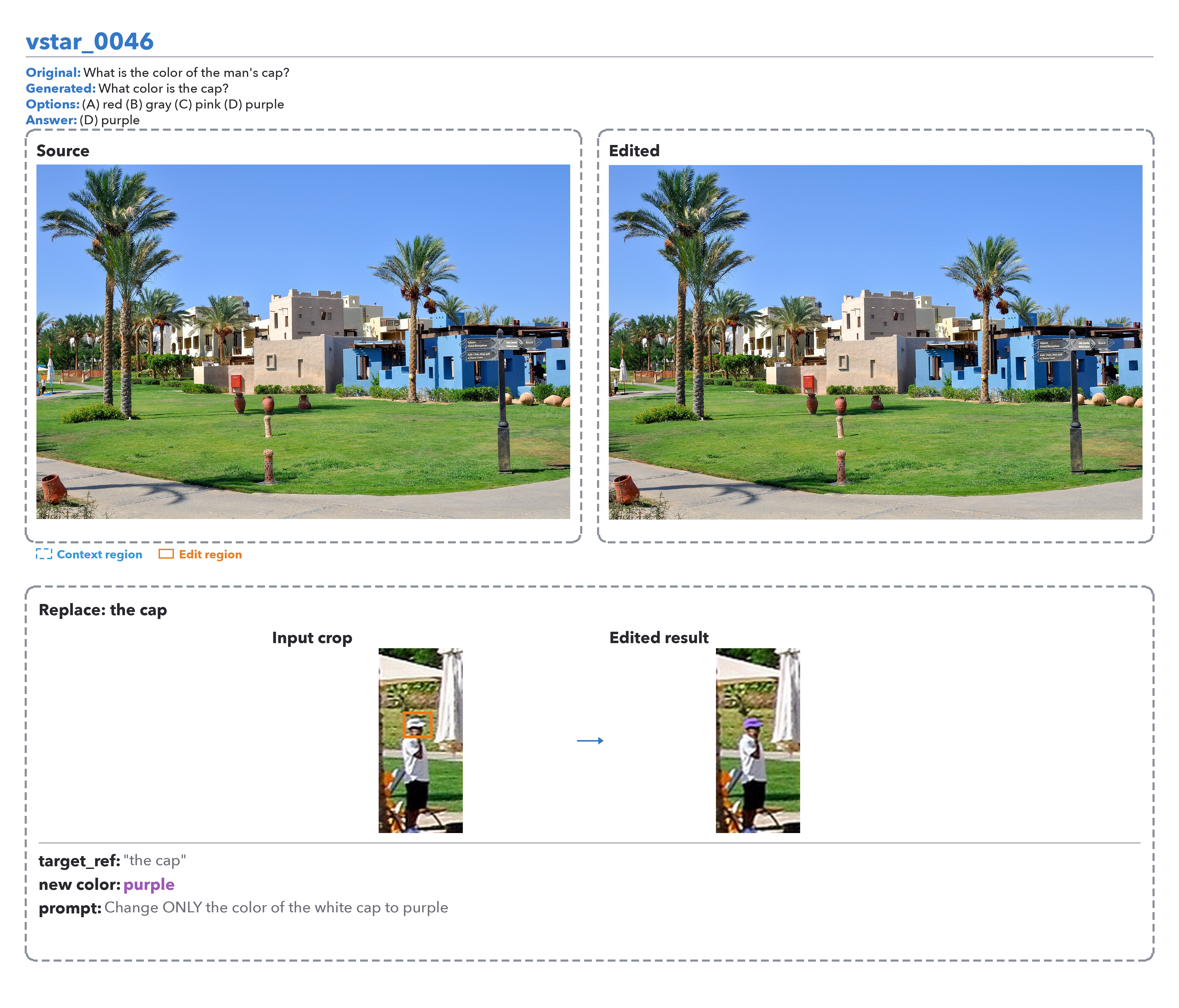}
  \caption{Success: replace mode (vstar\_0046). A white cap in
  a resort scene is recolored to purple. Original accuracy: 88\%;
  \sysname{} accuracy: 38\%.}
  \label{fig:succ-replace}
\end{figure*}

\paragraph{Replace mode: color renewal on the original target.}
Figure~\ref{fig:succ-replace} shows vstar\_0046.
The original question asks about a man's cap color in a resort scene;
seven of eight models answer correctly (88\%).
\sysname{} recolors the cap from white to purple.
The object's identity and location are unchanged, so the visual-search
task is largely preserved; yet only three models identify the new color
(38\%).
This is consistent with the original high accuracy reflecting a
memorized answer rather than genuine localization of the cap.

Together, these two cases illustrate why both slot modes contribute to
decontamination: add mode eliminates scene-object associations, and
replace mode invalidates memorized attributes, while both preserve the
small-target, cluttered-scene regime that defines V*Bench's difficulty.